%% file: main.tex
\documentclass[conference]{IEEEtran}
\IEEEoverridecommandlockouts
\usepackage{cite}
\usepackage{amsmath,amssymb,amsfonts, dsfont}
\usepackage{algorithmic}
\usepackage{graphicx}
\usepackage{textcomp}
\usepackage{xcolor}
\usepackage{hyperref}
\usepackage{url}
\usepackage{enumitem}
\usepackage{dsfont}
\usepackage{amssymb}
\usepackage{subcaption}
\usepackage[square, sort, comma, numbers]{natbib}

\usepackage[linesnumbered, ruled]{algorithm2e}
\SetKwComment{Comment}{$\triangleright$\ }{}

\newtheorem{theorem}{Theorem}

\newtheorem{proposition}{Proposition}
\newtheorem{definition}{Definition}
\newtheorem{lemma}{Lemma}

\newtheorem{corollary}{Corollary}
\newenvironment{proof}{{\it{Proof:}}}{\hfill$\square$}
\DeclareMathOperator*{\argmin}{arg\,min}
\input{edits}

\def\BibTeX{{\rm B\kern-.05em{\sc i\kern-.025em b}\kern-.08em
    T\kern-.1667em\lower.7ex\hbox{E}\kern-.125emX}}
\begin{document}

\title{Learning safety critics via a non-contractive\\binary Bellman operator
\thanks{This work was supported by NSF through grants CAREER 1752362, CPS 2136324, Global Centers 2330450.}
}

\author{\IEEEauthorblockN{Agustin Castellano}
\IEEEauthorblockA{\textit{Johns Hopkins University} \\
Baltimore, MD, USA \\
\texttt{acaste11@jhu.edu}}
\and
\IEEEauthorblockN{Hancheng Min}
\IEEEauthorblockA{\textit{University of Pennsylvania} \\
Philadelphia, PA, USA \\
\texttt{hanchmin@seas.upenn.edu}}
\and
\IEEEauthorblockN{Juan Andr\'es Bazerque}
\IEEEauthorblockA{\textit{University of Pittsburgh} \\
Pittsburgh, PA, USA \\
\texttt{juanbazerque@pitt.edu}}
\and
\IEEEauthorblockN{Enrique Mallada}
\IEEEauthorblockA{\textit{Johns Hopkins University} \\
Baltimore, MD, USA \\
\texttt{mallada@jhu.edu}}
}

\maketitle

\begin{abstract}
The inability to naturally enforce safety in Reinforcement Learning (RL), with limited failures, is a core challenge impeding its use in real-world applications. One notion of safety of vast practical relevance is the ability to avoid (unsafe) regions of the state space. Though such a safety goal can be captured by an action-value-like function, a.k.a. safety critics, the associated operator lacks the desired contraction and uniqueness properties that the classical Bellman operator enjoys. In this work, we overcome the non-contractiveness of safety critic operators by leveraging that safety is a binary property. To that end, we study the properties of the binary safety critic associated with a deterministic dynamical system that seeks to avoid reaching an unsafe region. We formulate the corresponding binary Bellman equation (B2E) for safety and study its properties. While the resulting operator is still non-contractive, we fully characterize its fixed points representing--except for a spurious solution--maximal persistently safe regions of the state space that can always avoid failure. We provide an algorithm that, by design, leverages axiomatic knowledge of safe data to avoid spurious fixed points. 
\end{abstract}


\section{Introduction}

The last decade has witnessed a resurgence of Reinforcement Learning (RL) as a core enabler of Artificial Intelligence (AI).
Today, RL algorithms can provide astonishing demonstrations of super-human performance in multiple settings, such as Atari~\cite{mnih2015human}, Go~\cite{silver2016mastering}, StarCraft II~\cite{vinyals2017starcraft}, and even poker~\cite{nichols2019machine}. However, this super-human success in RL is overwhelmingly limited to \emph{virtual domains} (games in particular), where not only one has a vast amount of data and computational power, but also there is little consequence to failure in achieving a task. Unfortunately, physical domain applications (autonomous driving, robotics, personalized medicine) lack most of these qualities and are particularly sensitive to scenarios where the consequences of poor decision-making are catastrophic \cite{yu2021reinforcement},\cite{brunke2022safe}.

Guaranteeing safety in an RL setting is challenging, as agents often lack a priori knowledge of the safety of states and actions~\cite {gu2022review}. Inspired by these challenges, numerous methods have been proposed to imbue RL methods with safety constraints, including expectation constraints \cite{paternain2022safe,castellano2023learning}, probabilistic/conditional value at risk constraints \cite{chow2017risk,chen2023probabilistic}, and stability constraints \cite{li2020robust, taylor2020learning}, among others. Such methods naturally lead to different safety guarantees, some of which can be theoretically characterized~\cite{robey2020learning,castellano2022reinforcement}. {\color{black}However, most of these methods fail to capture the safety-critical nature of some events that must be avoided at all costs, i.e., with probability one.}

One type of safety constraint of practical relevance in safety-critical applications is reachability constraints (e.g.\cite{bertsekas1972infinite}; \cite[Ch. 3]{sontag2013mathematical}; \cite{bansal2017hamilton}), wherein one seeks to avoid regions of the state space that are associated with failure events by computing sets that are either, persistently safe (a.k.a. control invariant safe sets~\cite{gurriet2018towards}), i.e., regions of the state that can avoid failure regions \emph{for all times} by proper choice of actions, or unsafe regions (a.k.a. as backward reachable tubes~\cite{mitchell2007comparing}) where \emph{failure is unavoidable} irrespectively of the actions taken. Recent research efforts incorporating such constraints in RL problems have shaped the notion of safety critics~\cite{fisac2019bridging, srinivasan2020learning, thananjeyan2021recovery}, which aim to compute action value-like functions that, based on information about either the (signed) distance to failure or a logical fail/not fail feedback, predict whether a state-action pair is safe to take or is doomed to catastrophic failure. 

Unfortunately, the computation and learning of safety critics is a challenging task since their corresponding Bellman-like equations (and associated operators) lack typical uniqueness (resp. contraction) properties that guarantee the validity of the solution (and convergence of RL algorithms). As a result, most works seek to compute approximate safety critics by introducing an artificial discount factor~\cite{fisac2019bridging, hsu2021safety}. This approximation, however, can drastically affect the accuracy of the critic, as approximately safe sets are not, by design, safe.

\enrique{I think we need to emphasize the last point I make here. That is, approximate safe sets/control invariant sets are not  safe/invairiant.}

\paragraph{Contributions of our work} 
In this work, we seek to overcome the difficulties in computing accurate safety critics by developing supporting theory and algorithms that allow us to learn accurate safety critics directly from the original non-contractive safety critic operator. Precisely, we consider a setting with deterministic, continuous state dynamics that are driven by stochastic policies on discrete action spaces, and model safety as a binary (\texttt{safe} / \texttt{unsafe})  quantity. Building on the literature of risk-based safety critics, we develop a deeper theoretical understanding of the properties of the corresponding \emph{binary safety (action-)value function} and how to exploit them to learn accurate safety critics. In doing so, we make the following contributions.

\begin{itemize}[leftmargin=0.5cm]
    \item \textbf{Characterization of solutions to the binary Bellman equations for safety} 
    We study the properties of the \emph{action-value function} associated with the binary safe/unsafe feedback and formulate a \emph{binary Bellman equation} (B2E) that such function must satisfy.  This B2E is undiscounted and has a non-contractive operator with multiple fixed points. Nevertheless, we show (Theorem \ref{thm:fixed-points}) that all (but one) of the possibly infinite solutions to the B2E represent regions of the state space that are: \emph{(i)} persistently safe regions that can avoid failure for all future times and \emph{(ii)} maximal, in the sense that no state that is declared to be unsafe can reach the declared safe region.


    \item \textbf{Algorithm for learning fixed points of a non-contractive operator} Finally, we provide an algorithm that can find a fixed point of the non-contractive operator, despite the lack of contraction. Our algorithm has two distinctive features that make this possible. First, it uses \emph{axiomatic data points}, i.e., points of the state space that are a priori known to be safe. Secondly, it uses a classification loss 
    that enforces \emph{self-consistency} of the Bellman equation across samples.
    \textcolor{black}{Preliminary numerical evaluations indicate that our proposed methodology outperforms a well-known safety critic \cite{fisac2019bridging} in a simple setup.}
    
\end{itemize}

\section{Problem Formulation}\label{sec:background}
\paragraph{Environment} We consider a Markov Decision Process $\langle\mathcal{S}, \mathcal{A}, F, \mathcal{G}, i, \rho\rangle$ where the state space $\mathcal{S}$ is \emph{continuous} and compact, the action space $\mathcal{A}$ is \emph{discrete} and finite, the map $F:\mathcal{S}\times\mathcal{A}\rightarrow\mathcal{S}$ is a \textit{deterministic} transition function. The set $\mathcal{G}$ represents a set of ``failure'' states to be avoided. At each time step, the agent receives as feedback the \emph{insecurity} of state $s_t$, that is  
$
i(s_t) = \mathds{1}\{s_t\in\mathcal{G}\}\in\{0,1\}.
$
Episodes start at a state $s_0\sim\rho$ and run indefinitely or end when the system enters $\mathcal{G}$.

\paragraph{Policies} We consider stochastic, stationary policies $\pi:\mathcal{S}\rightarrow\Delta_{\mathcal{A}}$ in the simplex $\Delta_{\mathcal{A}}$, and denote $\pi(a|s)$ the probability of  $a\in\mathcal{A}$ when at state $s\in\mathcal{S}$. With discrete and finite $\mathcal{A}$ and deterministic transition dynamics, the set of reachable states starting from any $s$ is finite, as defined next

\begin{definition}[$t$-step reachable sets]
    For any policy $\pi$ and any state $s\in\mathcal{S}$, the $t$-step reachable set from $s$ under $\pi$ is
        $\mathcal{F}^\pi_t(s) \triangleq \big\{s'\in\mathcal{S}: \mathbb{P}^\pi\left(s_t=s'\mid s_0=s\right) > 0 \big\}.$
    Similarly, for any $a\in\mathcal{A}$ we define 
    $\mathcal{F}^\pi_t(s,a) \triangleq \big\{s'\in\mathcal{S}: \mathbb{P}^\pi\left(s_t=s'\mid s_0=s, a_0=a\right) > 0 \big\}.$
 
\end{definition}

\begin{figure}[t]
    \centering
    \includegraphics[width=.7\linewidth]{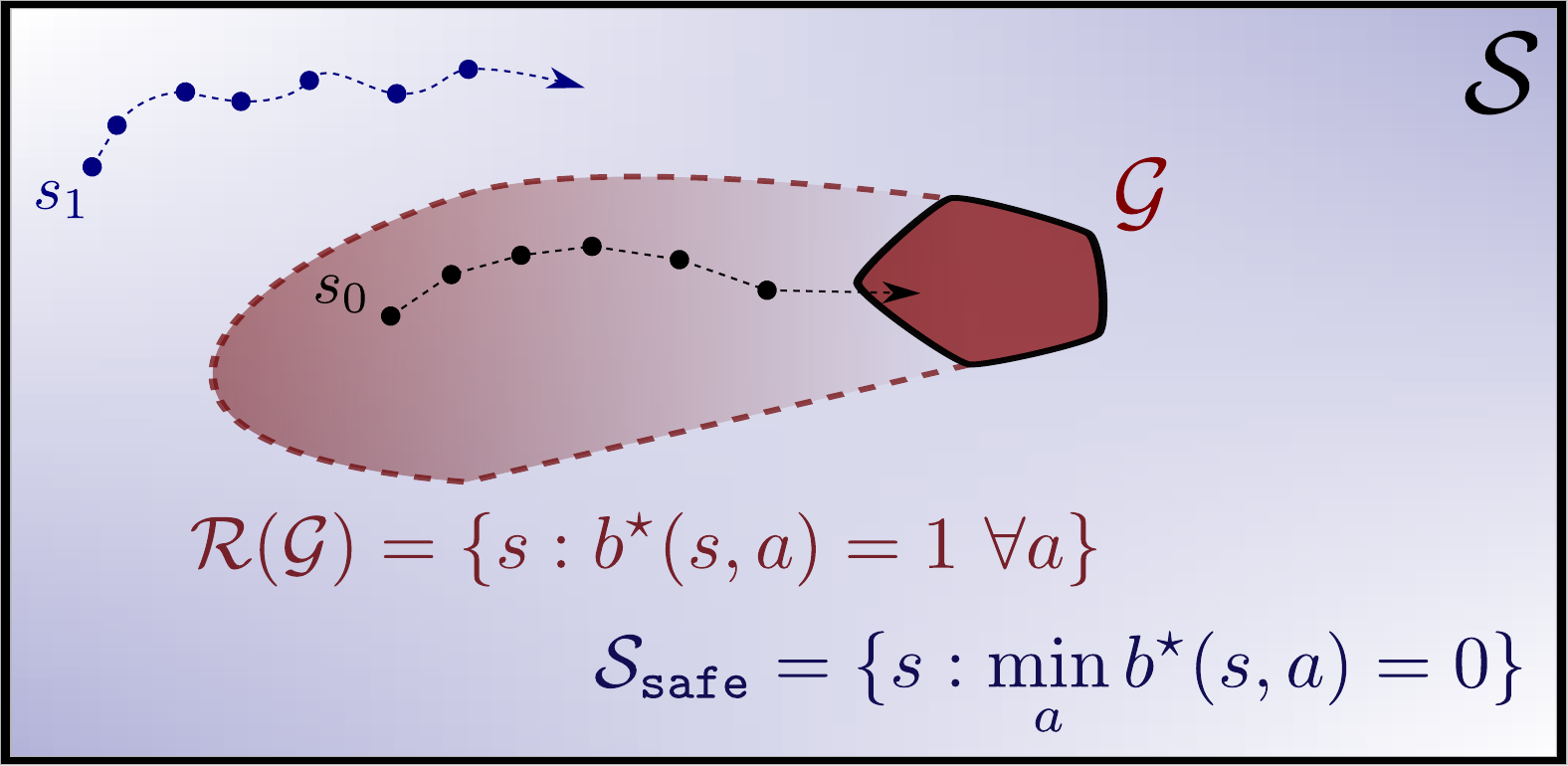}
    \caption{The optimal $b^\star$ describes different regions of the state space. The set $\mathcal{G}$ (solid red) is to be avoided at all times. Due to system dynamics, there is a region of the state space $\mathcal{R}(\mathcal{G})$ (shaded red) such that any trajectory starting there (e.g., from $s_0$) will inevitably enter $\mathcal{G}$. For any point in its complement $\mathcal{S}_{\texttt{safe}}$ (e.g. $s_1$), the optimal policy avoids $\mathcal{G}$ at all times.}
    \label{fig:safe-set}
\end{figure}

Given these notions of reachable sets, we can define the binary safety value functions for any policy.
\begin{definition}[Binary safety value functions] 
    {The binary safety (action-)value function of policy $\pi$ at state $s$ (and action $a$) is:
    \begin{align}
        v^\pi(s) &\triangleq \sup_{t\geq 0}\max_{s_t\in \mathcal{F}^\pi_t(s)} i(s_t)\,,\\
        b^\pi(s, a)&\triangleq \sup_{t\geq 0}\max_{s_t\in \mathcal{F}^\pi_t(s, a)} i(s_t)\,. \label{eq:safety-action-value-function}
    \end{align}}
\end{definition}

We choose the notation $b(\cdot,\cdot)$ instead of the usual $Q$ to emphasize that it is a binary action-value function. Note that $b^\pi(s, a)=1$ if and only if starting from $(s, a)$ and following $\pi$, there is a positive probability of entering $\mathcal{G}$. The optimal (action-)value functions are then defined.
\begin{definition}[Optimal binary value functions] For all $s\in\mathcal{S}$ and $a\in\mathcal{A}$, the optimal value and action-value functions are
    $v^\star(s)\triangleq \min_{\pi}v^\pi(s)$ 
    and $b^\star(s, a)\triangleq \min_{\pi}b^\pi(s,a).$
\end{definition}
\paragraph{Relationship between safety and the optimal binary functions} These optimal value functions fully characterize the logical \texttt{safe/unsafe} nature of each state or state-action pair, and have nice interpretations in terms of how they partition the state-space, as illustrated in Fig. \ref{fig:safe-set}. Recall that the safety goal is to avoid $\mathcal{G}$. However, due to the MDP dynamics, this might not be possible for every state outside $\mathcal{G}$\footnote{A car heading to a wall ($\mathcal{G}$) one meter away at 100mph will hit it, regardless of the actions taken.}. A state $s$ is persistently safe if trajectories from $s$ can avoid $\mathcal{G}$ at all times---in other words, if $\exists a\in\mathcal{A} : b^\star(s,a)= 0$. Conversely, a state $s$ is doomed to fail if $b^\star(s,a) =1~\forall a\in\mathcal{A}$. We use $\mathcal{R}(\mathcal{G})$ to denote this set of ``unsafe states'' that are doomed to fail.
The complement of this set is the set of persistently safe states, and the ``safe'' actions for each state are given by:
\begin{align}
    &\mathcal{S}_{\texttt{safe}} = \big\{s\in\mathcal{S}: \min_{a\in\mathcal{A}}b^\star(s, a)=0\big\} \label{eq:s-safe}\\
    &\mathcal{A}_{\texttt{safe}}(s) = \big\{a\in\mathcal{A}: b^\star(s,a) = 0\big\}
 .
\end{align}
Just like in the standard RL setup, each (action-)value function has associated Bellman equations.

\begin{proposition}[Binary Bellman Equations]\label{prop:binary-bellman-equations}
For any policy $\pi$, the following set of Bellman equations hold for all $s\in\mathcal{S}$, for all $a\in\mathcal{A}$: $b^\pi(s,a) = i(s) + \big(1-i(s)\big)v^\pi(s')$, where $s'=F(s,a)$.
In particular, any optimal policy satisfies: 
\begin{equation} b^\star(s,a) = i(s) + \big(1-i(s)\big)\min_{a'\in\mathcal{A}}b^\star(s',a').   \label{eq:bellman-b}
    \end{equation}
\begin{proof}
   See Appendix \ref{ap:proof-of-bbe}.
\end{proof}
\end{proposition}
\paragraph{Unsafety as a logical \texttt{OR}}
The Bellman equation for the optimal $b^\star$ can be understood as: ``an $(s,a)$ pair is \emph{unsafe} ($b^\star(s,a)=1$) if either: the current state is unsafe ($i(s)=1$), \texttt{OR} it leads to an unsafe state later in the future ($\min_{a'}b^\star(s',a')=1$).''
\paragraph{Non-contractive Bellman operator}
The optimal binary function of \eqref{eq:bellman-b} has an associated operator, acting on the space of functions $\mathcal{B}=\left\{b:\mathcal{S}\times\mathcal{A}\to\{0, 1\}\right\}$, $\mathcal{T}:\mathcal{B}\to\mathcal{B}$ s.t.
\begin{equation}
    (\mathcal{T}b)(s,a) = i(s) + (1-i(s))\min_{a'\in\mathcal{A}}b(s',a')\quad\forall (s,a) \label{eq:b-operator}
\end{equation}
One of the key features in the standard (discounted) Bellman equations for infinite-horizon problems is that it has an associated operator that is contractive \cite[p.45]{bertsekas2015dynamic}. As such, it admits a unique fixed point (the optimal value function). This is crucial for applying value iteration procedures or for methods reliant on temporal differences \cite{schwartz93}. Surprisingly, the operator defined in \eqref{eq:b-operator} is non-contractive, and as such, it admits more fixed points than the optimal $b^\star$.
In the next section we will see that all but one of these fixed points possess the desired safety properties.
\juan{I don't understand this sentence. It's sounds contradictory.}
\subsection{Closely Related Work}\label{sec:closely-related}
\paragraph{Control-theoretic approaches for computing $\mathcal{S}_{\texttt{safe}}$} Standard tools from Control Theory exist to approximate the safe regions corresponding to $b^\star(\cdot,\cdot)$, both for linear  \cite{girard2006efficient} and non-linear dynamics \cite{mitchell2005toolbox}. The latter requires knowledge of the transition map $F(\cdot,\cdot)$ along with the signed distance to the unsafe region \cite{mitchell2005time}. This amounts to solving partial differential equations (PDEs) of the Hamilton-Jacobi-Isaacs (HJI) type \cite{bansal2017hamilton}, and yields value functions whose zero super-level sets correspond to $\mathcal{S}_{\texttt{safe}}$. 
\paragraph{Risk-based vs Reachability-based safety critics}
The binary action-value function $b^\star$ defined in \eqref{eq:bellman-b} is closely related to recent work on \emph{Risk-based} safety critics~\cite{srinivasan2020learning,thananjeyan2021recovery}, which use binary information to indicate the risk of unsafe events. However, unlike risk-based critics, which seek to measure a cumulative expected risk $b^\star_\text{risk}(s,a)=\max_{\pi}E_\pi[\sum_{k=t}^\infty\gamma ^{k-t} i(s_t)| s_t=s, a_t=a]\in[0,1]$, our binary critic only takes values  $b^*(s,a)\in\{0,1\}$, and outputting $1$ whenever unavoidable failure has positive probability. 
\emph{Reachability based} safety critics, build on the literature of HJI equations and seek to measure the largest (signed) distance $h(s_t)$ that one can sustain from the failure set $\mathcal{G}$, i.e., $b^\star_\text{reach}(s,a)=\sup_{\pi}\inf_{t\geq0}h(s_t)$ almost surely ~\cite{fisac2019bridging}. Our binary critic $b^\star$ is indeed related to $b^\star_\text{reach}$ when the signed distance $h(s)$ is replaced with the binary signal $-i(s)$. We will soon show that this particular choice of safety measure allows for a precise characterization of the fixed points of \eqref{eq:b-operator}.
\paragraph{To contract or not to contract}
Despite the diversity of safety critics present in the literature, a common practice in both risk-based critics~\cite{srinivasan2020learning,thananjeyan2021recovery} and reachability-based critics~\cite{fisac2019bridging,chen2021safe}
is the introduction of a discount factor $\gamma<1$. {\color{black}While this leads to desired uniqueness and contraction properties for the operator, it comes with trade-offs: it degrades the accuracy, requiring the introduction of conservative thresholds \cite{srinivasan2020learning,chen2021safe}, which further limits exploration. Notably}, such an approach is particularly problematic when seeking to guarantee persistent safety (the ability to avoid failure for all future times), as such property is not preserved for finite accuracy approximation, even for thresholded ones. In this work, we overcome this limitation by seeking to learn directly using the non-contractive operator, thus guaranteeing, by design, the correctness of the solution.

\section{Binary characterization of safety}\label{sec:binary}

The fixed points $b^\star$ of the binary Bellman operator have a meaningful interpretation in terms of the topology of the state-space and can be used to derive persistently safe policies. This connection will be better understood once we define the notion of control invariant safe sets.

\begin{definition}[Control invariant safe (CIS) set]\label{def:cis}
    A set $\mathcal{C}\subset\mathcal{S}$ is a control invariant safe (CIS) set  if there exists a policy $\pi$ such that:
    \begin{enumerate}[label=\roman*)]
        \item (Control invariance): $\forall s_0\in\mathcal{C}, \forall t\geq 0, \quad\mathcal{F}_t^\pi(s_0)\subset\mathcal{C}$
        \item (Safety): $\quad\quad\quad\quad\quad\forall s_0\in\mathcal{C}, \forall t\geq 0, \quad\mathcal{F}^\pi_t(s_0) \cap \mathcal{G} = \emptyset.$
    \end{enumerate}
\end{definition}
In essence, \textit{(i)} means that there exists a controller that guarantees that trajectories starting in $\mathcal{C}$ can be made to remain in $\mathcal{C}$ forever, which is a standard notion in control theory \cite{bertsekas1972infinite, blanchini1999set}. Property \textit{(ii)} means this can be done while also avoiding the unsafe region $\mathcal{G}$!

\begin{figure}
    \centering
    \includegraphics[width=\linewidth]{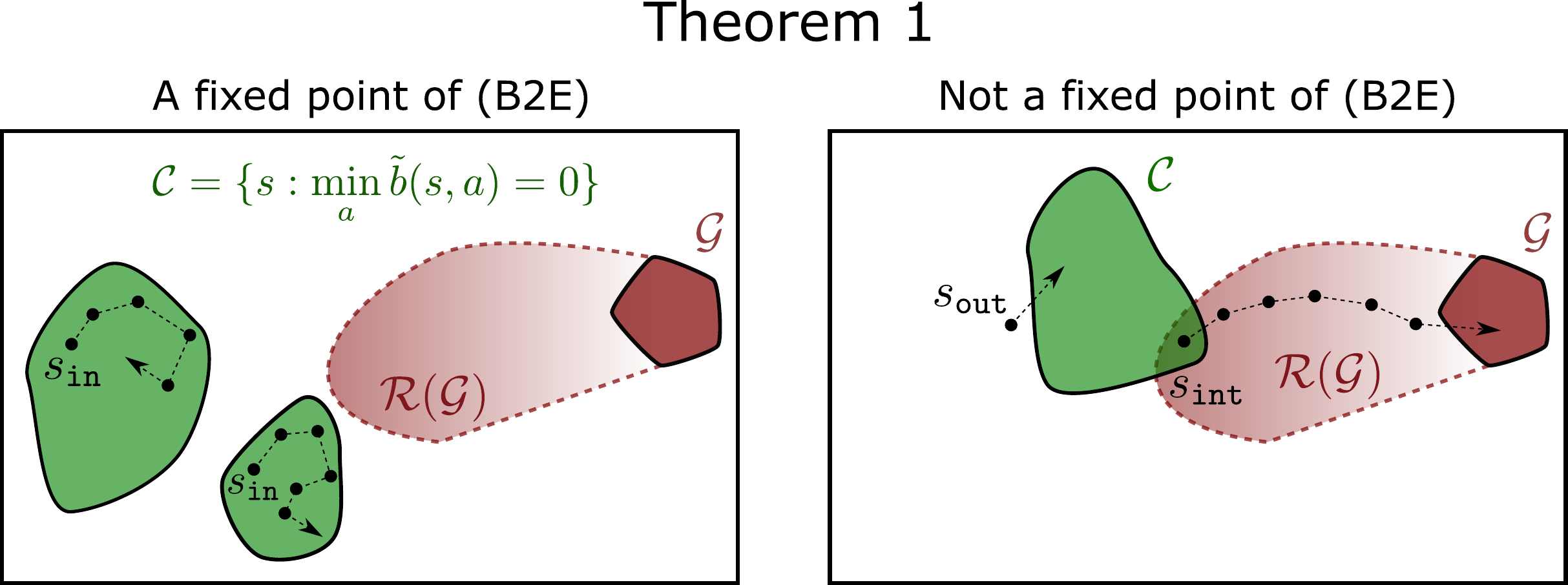}
    \caption{An illustration of Theorem \ref{thm:fixed-points}. Left: a valid fixed point $\tilde{b}$ of $\mathcal{T}$ and its corresponding safe control invariant set. Trajectories starting in $\mathcal{C}$ can be driven to remain in $\mathcal{C}$. Right: a function $\tilde{b}$ that is not a fixed point. A state $s_{\texttt{int}}$ in the intersection will inevitably lead to the unsafe region $\mathcal{G}$, so $\tilde{b}(s,a)$ should be $1$ for all states in the trajectory (which would mean $s_\texttt{int}\notin\mathcal{C}$). Similarly, a state $s_{\texttt{out}}$ outside $\mathcal{C}$ cannot reach inside. If it could, $\tilde{b}(s_{\texttt{out}},a)=1$ for some $a\in\mathcal{A}$,  but it would transition to a state where $\min_{a'}\tilde{b}(s',a')=0$, violating \eqref{eq:bellman-b}. }
    \label{fig:fixed-points} 
\end{figure}

With these definitions in place, we are ready for our main result.

\begin{theorem}[Fixed points and control invariant safe sets]\label{thm:fixed-points}
    Let $\tilde{b}:\mathcal{S}\times\mathcal{A}\rightarrow\{0, 1\}$ be a fixed point of \eqref{eq:b-operator}. Then either $\tilde{b}(s,a)=1~\forall (s,a)$ (spurious fixed point), or:
    \begin{enumerate}[label=\roman*)]
        \item $\mathcal{C} \triangleq \big\{s\in\mathcal{S}: \min_{a}\tilde{b}(s,a) = 0\big\}$ is control invariant safe (CIS).  
        \item $\mathcal{C}$ is unreachable from outside, i.e.,
        $\mathcal{F}_t^\pi(s_0) \cap \mathcal{C} = \emptyset
            \quad\quad\forall s_0\in\mathcal{S}\setminus\mathcal{C}, \forall \pi, \forall t\geq 0.$
        \item Any policy $\pi$ that satisfies \eqref{eq:charact-safe-policies} renders $\mathcal{C}$ CIS.
    \begin{equation}
         \tilde b(s,a)=1\ \Rightarrow \pi(a|s)=0,\ \forall s\in\mathcal C.\label{eq:charact-safe-policies}
    \end{equation}
    \end{enumerate}
\begin{proof}
The proof is in Appendix \ref{ap:proof-of-fixed-points}.
\end{proof}
\end{theorem}
The first statement proclaims that starting in $\mathcal{C}$, the system can be made to remain in $\mathcal{C}$ forever (thus ensuring safety). The contrapositive of property \emph{(ii)} sheds light on a notion of \emph{maximality} of $\mathcal{C}$:
\begin{corollary}[Maximality of the CIS set]\label{cor:maximal}
     Let $\mathcal{X}$ be a strict subset of $\mathcal{C}$. If $\mathcal{X}$ is reachable\footnote{i.e. if $\exists \pi, \exists t\geq 0, \exists s_0\in\mathcal{C}\setminus\mathcal{X}: \mathcal{F}^\pi_t\left(s_0\right)\cap\mathcal{X}\neq\emptyset$} from $\mathcal{C}\setminus\mathcal{X}$, 
    then $\mathcal{X}$ cannot be associated\footnote{that is to say: $\forall \tilde{b}: \tilde{b} = \mathcal{T}\tilde{b}, \mathcal{X}\neq  \big\{s\in\mathcal{S}: \min_{a}\tilde{b}(s,a) = 0\big\}$} with any fixed point of \eqref{eq:b-operator}.
    
\end{corollary}

We refer the reader to Fig. \ref{fig:fixed-points} for an illustration of valid and nonvalid fixed points.
By means of Theorem \ref{thm:fixed-points} and Corollary \ref{cor:maximal}, we achieve our goal of identifying the fixed points of the binary Bellman operator to maximal persistently safe states. In the following section, we will present an algorithm that finds fixed points of $\mathcal{T}$.
%

\begin{algorithm}
\caption{Pseudocode for learning the binary value function }
\label{alg:main}
\KwIn{Safe dataset $\mathcal{D}_{\texttt{safe}}$;}
\KwOut{Binary value function~$b^\theta(\cdot,\cdot)$;}
Initialize $b^\theta(\cdot,\cdot)$ using $\mathcal{D}_{\texttt{safe}}$ and 
 $\mathcal{M}=[~]$ \Comment*{Transition buffer.}
\Repeat{termination}{

\For{i=0,\ldots\texttt{NUM\_EPISODES}-1}{
Run episodes, store $\big(s_k,a_k,i(s_k),s_k'\big)_{k=1}^K$ transitions in $\mathcal{M}$;
}

$\mathcal{D}_{\texttt{unsafe}}\leftarrow\texttt{build\_unsafe\_dataset}(b^\theta, \mathcal{M})$ \Comment*{Use $b^\theta$ to compute labels.}

Build $\mathcal{D} = \mathcal{D}_\texttt{safe} \cup \mathcal{D}_\texttt{unsafe}$ \Comment*{Complete dataset.}
\Repeat{$\texttt{Accuracy}(b^\theta, \mathcal{D}) = 1$}{
Run gradient steps on $\mathcal{L}_{\texttt{train}}$ \Comment*{Update $b^\theta$}
}
$\mathcal{D}_{\texttt{unsafe}}\leftarrow\texttt{build\_unsafe\_dataset}(b_i, \mathcal{M})$ \Comment*{$b^\theta$ has changed w.r.t.\textbf{6}}
Build $\mathcal{D} = \mathcal{D}_\texttt{safe} \cup \mathcal{D}_\texttt{unsafe}$ \Comment*{New dataset}
\If{$\texttt{Accuracy}(b^\theta,\mathcal{D}) \neq 1$\Comment*{Check consistency of B2E}
} {\textbf{go to} Line \textbf{8} \Comment*{Not self-consistent $\Rightarrow$ Re-train the network}}
}
\end{algorithm}

\section{Algorithm}\label{sec:algo}

We propose training a neural network classifier to learn fixed points of $\mathcal{T}$ in \eqref{eq:b-operator}. We will denote the learned function by $b^\theta(\cdot,\cdot)$ where $\theta$ collects the network parameters. The network takes each state as input and outputs the value $b^\theta(s,a)$ for each possible action. The last layer is a point-wise sigmoid activation function ensuring $b^\theta(s,a)$ lies in the unit interval. We use $\hat{b}^\theta(s,a)\triangleq\texttt{Round}\left(b^\theta(s,a)\right)$ to denote the predicted label. Note that our threshold (at 1/2) will be fixed during training and testing.
The pseudocode for the main algorithm can be found in Alg. \ref{alg:main}. We provide a comprehensive breakdown of its main components next.
\paragraph{Dataset} The dataset $\mathcal{D}$ consists of $(s,a,y)$ tuples, where $y$ is a $\{0,1\}$ label, and has two components. A prescribed \footnote{e.g., $(s, a)$ pairs close to the system's equilibrium or sampled trajectories from a known, safe policy.} safe set $\mathcal{D}_{\texttt{safe}}$ (for which $y=0$) and a dynamically updated $\mathcal{D}_{\texttt{unsafe}}$ of unsafe transitions detected during data collection. We have observed empirically that the addition of $\mathcal{D}_{\texttt{safe}}$ helps prevent the resulting binary value function from collapsing into the trivial fixed point described in Theorem \ref{thm:fixed-points}. 
The algorithm iterates over the following three loops:
\paragraph{Environment interaction} Episodes start from a state $s_0$ sampled from the initial distribution $\rho$. To collect $(s,a,s',i(s))$ transitions and store them in a memory buffer $\mathcal{M}$ we run episodes by following a policy that satisfies \eqref{eq:charact-safe-policies}, for example the \emph{uniform safe policy}, which takes actions uniformly over the presumed-safe ones:
\begin{align}
    \pi^\theta(a|s) = \begin{cases}
        0&\text{~if~}\hat{b}^\theta(s,a)=1\\
        {1}/{\sum_{a'\in\mathcal{A}}\mathds{1}\{\hat{b}^\theta(s,a')=0\}}&\text{~if~}\hat{b}^\theta(s,a)=0
    \end{cases}\label{eq:uniform-safe-policy}
\end{align}
\paragraph{Building the dataset}
After collecting transitions, the binary value function is used to compute labels via the right-hand side of \eqref{eq:bellman-b}, that is,        $y_k^\theta = i(s)+(1-i(s))\min_{a'}{b}^\theta(s_k',a')$ for all $(s_k,a_k, i(s_k), s_k')\in\mathcal{M}$. 
Note that these are ``soft'' labels $y_k^\theta\in[0,1]$. Those that satisfy $y_k^\theta \geq \tfrac{1}{2}$ are added to $\mathcal{D}_{\texttt{unsafe}}$. This procedure is dubbed \texttt{build\_unsafe\_dataset($b,\mathcal M$)} in Algorithm \ref{alg:main}.
\paragraph{Training the network}
The network is trained by running mini-batch gradient descent on the binary cross-entropy loss until it can correctly predict all the labels in $\mathcal{D}:=\mathcal{D}_\texttt{safe}\cup\mathcal{D}_\texttt{unsafe}$. Once that is achieved, the labels in $\mathcal{D}_{\texttt{unsafe}}$ are re-computed (some might have changed since $b^\theta$ was updated during this process), and the program checks whether it can correctly predict the labels again. It repeats this process until all labels are predicted correctly, yielding a binary function that is self-consistent across the whole dataset.

\section{Numerical Experiments}\label{sec:experiments}
\begin{figure}[t]
    \centering
    \includegraphics[width=.4\linewidth]{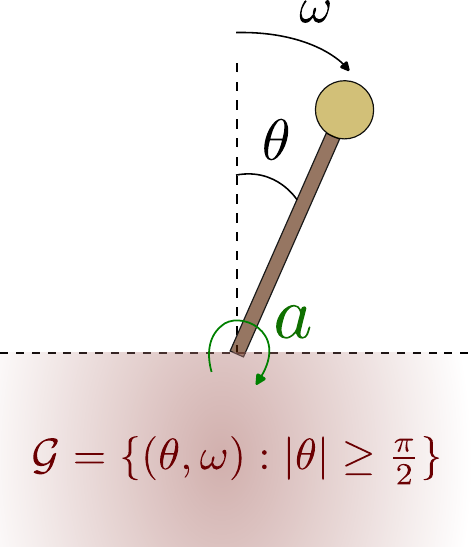}
    \caption{The custom inverted pendulum environment, with state $s=[\theta, \omega]^\top$. The region past the horizontal $\mathcal{G}$ is to be avoided at all times.}
    \label{fig:pendulum-mdp}
\end{figure}
\begin{figure}[t]
    \centering
    \includegraphics[width=\linewidth]{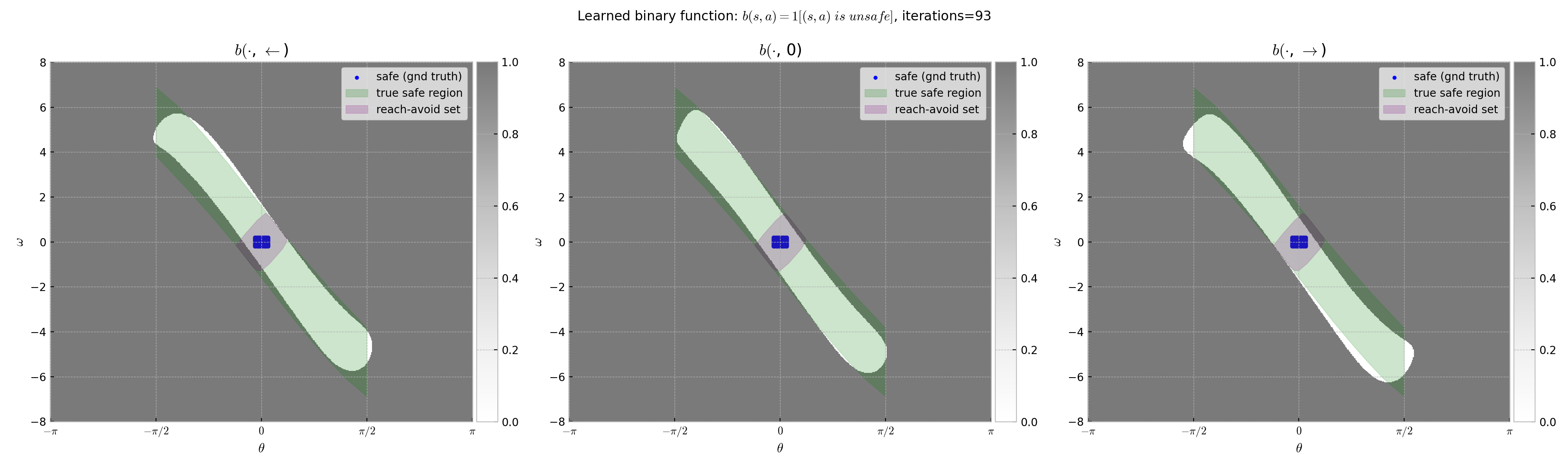}
    \caption{Learned safe regions for the inverted pendulum problem during training. Each panel depicts the learned barrier for a fixed action (maximum clockwise torque, maximum counter-clockwise torque, no torque). The white area corresponds to the states classified as safe (for each of those actions). The solid maroon lines show the boundary of the unsafe region $\mathcal{G}$ (falling past the horizontal). The green region shows the set of states that can avoid $\mathcal{G}$ at all times, and the purple region shows the set of safe states reachable from $\mathcal{D}_{\texttt{safe}}$. These sets were computed using an optimal control toolbox \cite{mitchell2005toolbox}.}
    \label{fig:pendulum-vf}
\end{figure}
We present numerical validations of our algorithm in an inverted pendulum environment \cite{towers_gymnasium_2023}, contrasting our method against SBE \cite{fisac2019bridging}, a well-known safety-critic. This environment allows easy visualization of the learned control invariant safe sets and can be compared against numerically obtained ``grounds truth'' references.
\paragraph{Environment} The state of the system $s=[\theta, ~\omega]^\top$ is the angular position and angular velocity of the pendulum with respect to the vertical. The action $a\in[-a_{\texttt{max}}, a_{\texttt{max}}]$ is the torque applied on the axis, which we discretize in $5$ equally spaced values. The \emph{goal} in this task is to avoid falling past the horizontal, i.e., $\mathcal{G} = \{ (\theta, \omega): |\theta| \geq \tfrac{\pi}{2}\}$, as show in Fig. \ref{fig:pendulum-mdp}. 

\paragraph{Training protocol} We take $\mathcal{D}_{\texttt{safe}}$ to be a small grid of $(s,a)$-pairs near the unstable equilibrium. Episodes are started from safe states (depicted in green in Fig. \ref{fig:pendulum-vf}) and end whenever the pendulum reaches the unsafe region, or after $200$ steps. The behavioral policy is the ``uniform safe'', as defined in \eqref{eq:uniform-safe-policy}. We alternate between collecting data for $10$ episodes, building the dataset, and training the network as explained in Sec. \ref{sec:algo}. Details on network architecture and hyperparameters are relegated to the Appendix \ref{ap:hyperparams}.
\paragraph{Ground truth} We compare the safe region learned by our algorithm against ground truths computed numerically with optimal control tools \cite{mitchell2005toolbox}.  Fig. \ref{fig:pendulum-vf} shows in green (resp. light gray) the maximum CIS set in the entire state (resp. the maximum CIS for trajectories that start inside the support of $\rho$).
The learned safe region (in white) at different stages of training is also shown. At the beginning, the network is only fit to $\mathcal{D}_{\texttt{safe}}$. As episodes run and it collects unsafe transitions, it effectively learns a CIS set included in the {true} safe region for the problem.

\begin{figure}
     \centering
     \begin{subfigure}[b]{0.45\linewidth}
         \centering
         \includegraphics[width=\linewidth]{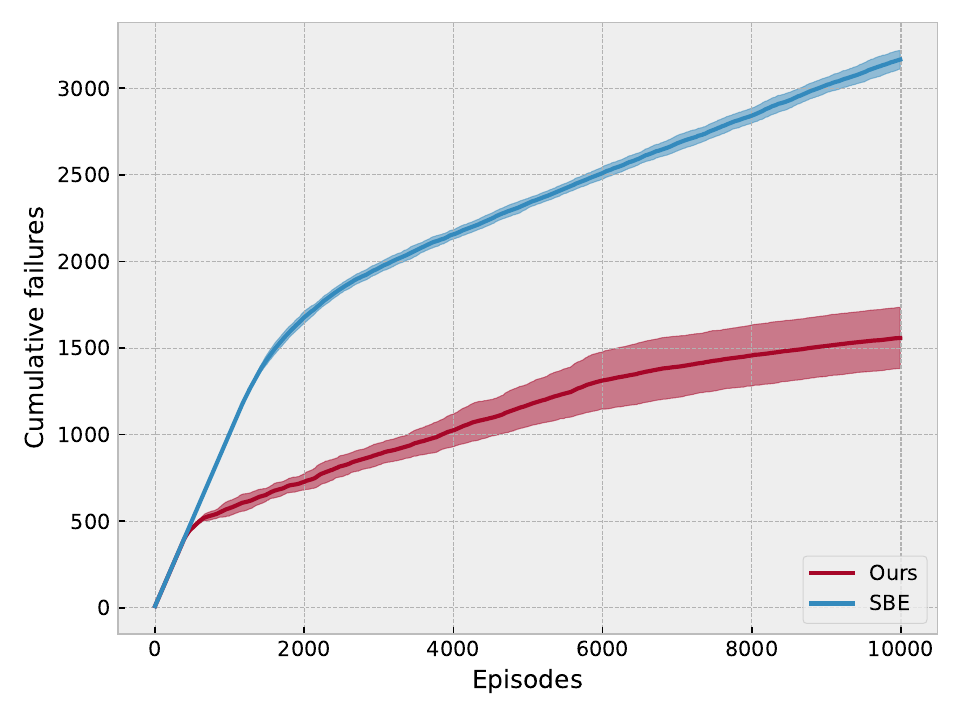}
     \end{subfigure}
     \begin{subfigure}[b]{0.45\linewidth}
         \centering
         \includegraphics[width=\linewidth]{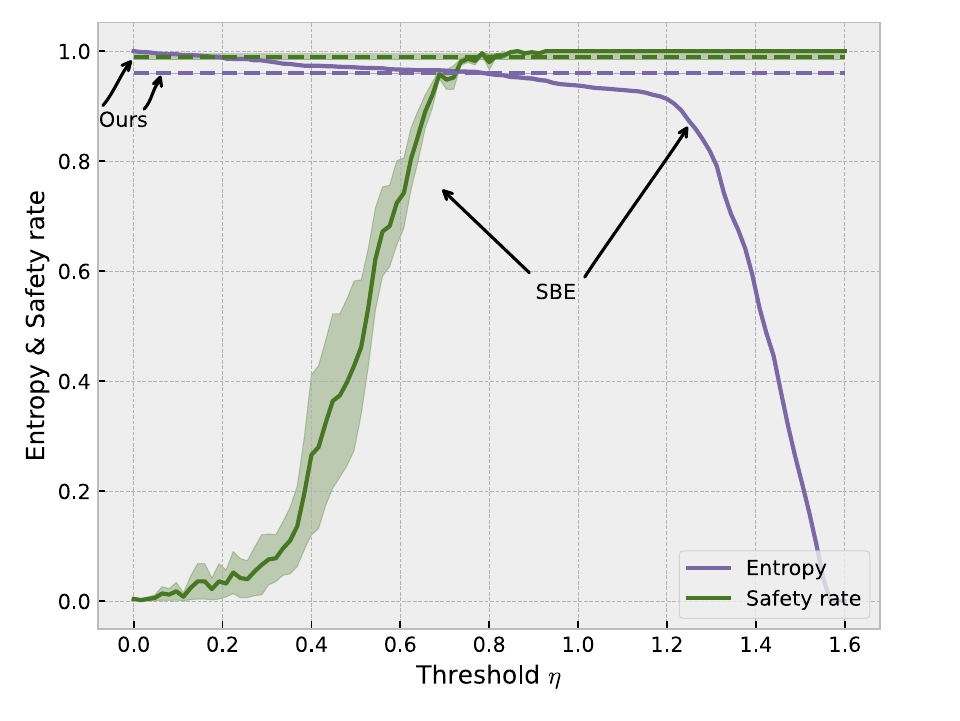}
     \end{subfigure}
     \hfill
     \caption{{Left: cumulative failures during training of our algorithm (red) and SBE (blue) for the inverted pendulum. Solid lines represent the means across $5$ seeds; shaded areas are $95\%$ confidence intervals. Our algorithm learns safe policies with less failures. Right: safety rate (fraction of safe episodes) and \emph{entropy} of each learned model.
     Our algorithm (shaded lines) always uses the uniform safe policy. SBE is tested for different threshold values $\eta$. Our policy achieves almost perfect safety rate and is exploratory (high entropy). Only the most conservative SBE policies (large $\eta$) are $100\%$ safe, but have low entropy (limited exploration).}}
    \label{fig:training-vs-dqn}
     \end{figure}

\paragraph{Training performance} We benchmark our proposed methodology against the Safety Bellman Equation (SBE) of \cite{fisac2019bridging}. This algorithm learns a safety-critic $q(s,a)$ and considers actions to be ``safe'' if $q(s,a)\geq\eta$ for a prescribed threshold $\eta$. 
Hyperparameters for that algorithm are taken from \cite{hsu2021safety} and can be seen in Appendix \ref{ap:hyperparams}. Fig. \ref{fig:training-vs-dqn} (left) shows the cumulative failures during training; (a \emph{failure} is an episode that touched the unsafe region $\mathcal{G}$). Our algorithm is significantly safer during training.

\paragraph{Post-training evaluation}
We evaluate the performance of each model after training and show it in Fig. \ref{fig:training-vs-dqn} (right). We test the uniform safe policy of our model against the safety critic for SBE. In the latter, we consider---for varying threshold $\eta$---the safe policy that maximizes exploration, i.e., the uniform policy taking actions $a$ such that $q(s,a)\geq\eta$. 
We illustrate the \emph{safety rate}, defined as the proportion of safe episodes, and the \emph{average entropy} of each policy $\tilde{\mathcal{H}}_\pi\triangleq\mathbb{E}_{s\sim\mathcal{RA}}\left[\mathcal{H}\left(\pi(\cdot\mid s)\right)\right]$, where $\mathcal{RA}$ is the set of safe states reachable from the origin (see `reach-avoid' set in Fig. \ref{fig:pendulum-vf}). Our algorithm obtains a perfect safety rate, while SBE only achieves it for safer policies (large enough $\eta$). These latter policies, though safe, are less exploratory---i.e., smaller entropy---than ours.
{\color{black}In summary, our achievements are twofold: we learn a \emph{persistently safe} family of policies that is \emph{more exploratory} than the SBE counterpart. As argued in Section \ref{sec:closely-related}, for traditional safety critics, there is no straightforward connection between the threshold $\eta$ and discount factor $\gamma<1$ needed to achieve safe policies, and safety comes at the expense of less exploration, which is undesired \textcolor{black}{and difficult to balance.} The solution found with our algorithm strikes a good balance between safety and the richness of the class of policies guaranteed to be safe.
}

\section{Conclusion}
We proposed a framework for obtaining correct-by-design safety critics in RL, to persistently avoid a region of the state space. Our framework exploits the logical safe/unsafe nature of the problem and yields binary Bellman equations with multiple fixed points. We argue that all these fixed points are meaningful by characterizing their structure in terms of safety and maximality. Numerical experiments validate our theory and show that we can safely learn safer, more exploratory policies.


\bibliography{refs}
\bibliographystyle{ieeetr}

\appendix
\section{Appendix}
\subsection{Some identities}
The $t$-step reachable set from $s$ is the union of the $(t-1)$-step reachable set from all the successor states of $s$:
\begin{equation}
    \mathcal{F}^\pi_t(s)=\bigcup_{s'\in \mathcal{F}^\pi(s)} \mathcal{F}^\pi_{t-1}(s') \label{eq:t-step-identity}
\end{equation}
Furthermore, the $t$-step reachable set from $(s,a)$ is the $t-1$ set from the successor $s'=F(s,a)$:
\begin{equation}\mathcal{F}^\pi_t(s,a) = \mathcal{F}^\pi_{t-1}(s') \quad\quad\forall t\geq 1\label{eq:t-step-tminus1}\end{equation}

\subsection{Proof of Proposition \ref{prop:binary-bellman-equations}}\label{ap:proof-of-bbe}
We will show:
$$b^\pi(s,a) = i(s) + \big(1-i(s)\big)v^\pi(s')\quad\quad\text{where~}s'=F(s,a). $$
The following identities hold, as explained below.
\begin{align}
    b^\pi(s,a) &= \sup_{t\geq 0}\max_{s_t\in\mathcal{F}^\pi_t(s,a)} i(s_t)\label{eq:pf-bbe-line1}\\
    &= \max\left\{i(s), ~\sup_{t\geq 1}\max_{s_t\in\mathcal{F}^\pi_t(s,a)} i(s_t)\right\}\label{eq:pf-bbe-line2}\\
    &= \max\left\{i(s), ~\sup_{t\geq 1}\max_{s_t\in\mathcal{F}^\pi_{t-1}(s')} i(s_t)\right\}\label{eq:pf-bbe-line3}\\
    &= \max\left\{i(s), ~\sup_{t\geq 0}\max_{s_t\in\mathcal{F}^\pi_{t}(s')} i(s_t)\right\}\label{eq:pf-bbe-line4}\\
    &= \max\left\{i(s), v^\pi(s')\right\}\label{eq:pf-bbe-line5}
\end{align}
The first identity is the definition of $b^\pi$. In \eqref{eq:pf-bbe-line2} unroll the first step in $\sup\{\cdot\}$. Next use the identity of \eqref{eq:t-step-tminus1}. Finally introduce the change of variables $t\gets t-1$ and recognize $v^\pi(s')$.

Recall $b^\pi(s,a), i(s)$ and $v^\pi(s')$ are binary. We consider two cases.

If $i(s)=1$:
\begin{equation}
    i(s)=1 \geq b^\pi(s,a) \geq i(s) \implies b^\pi(s,a) = i(s)
\end{equation}
If $i(s)=0$:
\begin{equation}
    b^\pi(s,a) = \max\{0, v^\pi(s')\} = 1\cdot v^\pi(s') = (1-i(s)) v^\pi(s')
\end{equation}
Hence:
$$
b^\pi(s,a) = i(s) + \big(1-i(s)\big)v^\pi(s'),
$$
which completes the first part of the proof.

Before proceeding to the last part of the proof, we note that a similar Binary Bellman equation holds for the value function (which we omitted in the manuscript for brevity):
\begin{proposition}[Binary Bellman equation for $v^\pi$]\label{prop:binary-bellman-equation-v} For all $\pi$, for all $s\in\mathcal{S}, a\in\mathcal{A}$:
    \begin{align*}
    v^\pi(s) = i(s) + \big(1-i(s)\big)\max_{s'\in\mathcal{F}^\pi(s)}v^\pi(s')
    \,;\quad s'=F(s,a)\,.
    \end{align*}
\begin{proof}
   We omit the proof since it is virtually identical to equations \ref{eq:pf-bbe-line1}--\ref{eq:pf-bbe-line5}.
\end{proof}
\end{proposition}

Now, going back to the proof of Proposition \ref{prop:binary-bellman-equations}, remains to be shown:
$$b^\star(s,a) = i(s) + \big(1-i(s)\big)\min_{a'\in\mathcal{A}}b^\star(s',a')~.
$$
In light of what we have just proved, it suffices to show the following Bellman optimality condition:
\begin{equation}
\min_{a\in\mathcal{A}}b^\star(s,a) = v^\star(s)~.\label{eq:b-star-v-star}
\end{equation}
We again consider two cases.

If $i(s)=1$, then $\forall \pi, \forall a\in\mathcal{A}$:
\begin{align*}
    1 =v^\pi(s) = b^\pi(s,a) \implies v^\star(s) = \min_a b^\star(s,a),
\end{align*}
so the result holds trivially.\\
If $i(s) = 0$:\\
Let $\pi^\star$ be an optimal policy. Then by Proposition \ref{prop:binary-bellman-equation-v}:
\begin{align}
    v^\star(s) &= \overbrace{i(s)}^{=0} + \overbrace{\big(1-i(s)\big)}^{=1}\max_{s'\in\mathcal{F}^{\pi^\star}(s)}v^\star(s')\\
    &= \max_{s'\in\mathcal{F}^{\pi^\star}(s)}v^\star(s') \\    &=\max_{a\in\supp\left[\pi(\cdot|s)\right]}v^\star(F(s,a))\\
    &\geq \min_{a\in\supp\left[\pi(\cdot|s)\right]}v^\star(F(s,a))\\
    &\geq \min_{a\in\mathcal{A}}v^\star(F(s,a))\\
    &=\min_{a\in\mathcal{A}}\left[i(s) + \big(1-i(s)\big)v^\star(F(s,a))\right] \\
    &=\min_{a\in\mathcal{A}}b^\star(s,a)
\end{align}

where the first inequality follows from $\max\geq \min$, and the second one for optimizing over a larger set.\\
Thus:
\begin{equation}
     v^\star(s) \geq \min_{a\in\mathcal{A}}b^\star(s,a)\quad\forall s\in\mathcal{S},\label{eq:prop1-wts}
\end{equation}
and we want to show that the result holds with equality. By contradiction, assume the inequality is strict for some $s\in\mathcal{S}$, that is to say: 
\begin{equation}
    \exists a^\dag\in\mathcal{A} : v^\star(s) > b^\star(s,a^\dag).
\end{equation}
Since the inequality is strict, it must be that $v^\star(s) = 1$ and $b^\star(s, a^\dag) = 0$.

Now consider a policy $\pi^\dag$ similar to $\pi^\star$, but that only takes action $a^\dag$ at state $s$:
\begin{equation}
    \pi^\dag:\begin{cases}
        \pi^\dag(a_1|s) = 1\\
        \pi^\dag(\cdot|s') = \pi^\star(\cdot|s') \quad\forall s'\neq s
    \end{cases}
\end{equation}
Let $v^\dag\triangleq v^{\pi^{\dag}}$. We then have:
\begin{align*}
    v^{\dag}(s)\!=\!\max_{s'\in\mathcal{F}^{\pi^\dag}(s)}v^{\dag}(s')\!=\!v^{\dag}(F(s, a^\dag))\!=\!b^\pi(s,a^\dag)\!<\!v^\star(s),
\end{align*}
hence 
$v^{\pi^\dag}(s) < v^\star(s)$ which means $\pi^\star$ is not optimal. This is a contradiction. It must then be that \eqref{eq:prop1-wts} holds with equality, as was claimed.

\subsection{Proof of Thm. \ref{thm:fixed-points}}\label{ap:proof-of-fixed-points}
\begin{proof}
Throughout this proof, we will make use of the following alternative representation of fixed points of the binary Bellman operator $\mathcal{T}$.
\begin{lemma} $\tilde{b}$ is a fixed point of $\mathcal{T}$ if and only if it satisfies, for all $s\in\mathcal{S}$, for all $a\in\mathcal{A}$:
\begin{equation} 
    \tilde{b}(s,a) = \max\left\{i(s), \min_{a'\in\mathcal{A}}\tilde{b}(s', a')\right\}\quad\text{where~}s'=F(s,a)~.
\end{equation}    
\begin{proof}
    The proof follows from equations \ref{eq:pf-bbe-line1}--\ref{eq:pf-bbe-line5} in Proposition \ref{prop:binary-bellman-equations} applied to $\pi^\star$.
\end{proof}
\end{lemma}

Now, to the main proof.
\paragraph{Spourious fixed point} Firstly, note that $\tilde{b}\equiv 1$ is indeed one possible fixed point of \eqref{eq:bellman-b}: $\forall (s,a)\in\mathcal{S}\times\mathcal{A},$
\begin{align*}
  1 = \tilde{b}(s,a) = i(s)+\big(1-i(s)\big)\min_{a'}\tilde{b}(s',a') \geq \min_{a'} \tilde{b}(s',a') = 1 
\end{align*}
\paragraph{$\mathcal{C}$ is CIS}Now suppose $\tilde{b}$ is non-trivial.
We begin by showing \textit{(i)}. By contradiction, assume $\mathcal{C}$ is not control invariant, i.e.:
$$
\forall \pi, \exists s_0\in\mathcal{C}, \exists t\geq 0 : \mathcal{F}^\pi_t(s_0)\not\subset\mathcal{C}. 
$$
We consider the ``safest'' policy that stems from $\tilde{b}$, only taking actions such that $\tilde{b}(s,a) = 0$. More generally, we consider any policy $\tilde{\pi}$ that satisfies:  
\begin{equation}
    \tilde{\pi}(s): \forall s\in\mathcal{S},\quad\supp\left[\tilde{\pi}(\cdot|s)\right]\subseteq \argmin_{a\in\mathcal{A}} \tilde{b}(s,a).
\label{eq:pi-safest-appendix}
\end{equation}
We have $\mathcal{F}_t^{\tilde{\pi}}(s_0) \not\subset\mathcal{C}$ for some $t\geq 0$. Hence there is a transition $(s,a,s')$ such that $s\in\mathcal{C}, a\in\argmin_{a'\in\mathcal{A}}\tilde{b}(s,a')$ and $s'=F(s,a)\notin\mathcal{C}$.
Therefore:
$$
0 \stackrel{(s\in\mathcal{C})}{=}  \tilde{b}({s},a) = \max\big\{i(s), \min_{a'}\tilde{b}(s',a')\big\} \geq \min_{a'}\tilde{b}(s',a')\stackrel{(s'\notin\mathcal{C})}{=}1,
$$
which is a contradiction. Hence $\mathcal{C}$ is control invariant---and moreover, the policy defined above renders it invariant (this shows \emph{(iii)}).\\
Now to show that $\mathcal{C}$ is safe, again assume by contradiction $\mathcal{C}$ is not safe, i.e.:
$$
\forall \pi, \exists s_0\in\mathcal{C}, \exists t\geq 0 : \mathcal{F}_t^\pi(s_0)\cap\mathcal{G} \neq \emptyset.
$$
Consider once again a ``safest'' policy as defined in \eqref{eq:pi-safest-appendix} (that renders $\mathcal{C}$ invariant). This policy along with the non-empty intersection in the previous equation implies that:
$$
\exists s \in\mathcal{C}, ~ a\in\supp\left[\pi(\cdot|s\right)],~ t\geq 0: s'=F(s,a)\in\mathcal{F}_t^\pi(s_0)\cap\mathcal{G} \implies$$
\begin{align*}
0 \stackrel{(s\in\mathcal{C})}{=}&\tilde{b}(s,a) = \max\big\{i(s), \min_{a'\in\mathcal{A}}\tilde{b}(s',a')\big\}\\
\geq& \min_{a'\in\mathcal{A}}\tilde{b}(s',a') = \min_{a'\in\mathcal{A}} \max\big\{i(s'), \min_{a''\in\mathcal{A}}\tilde{b}\left(F(s',a'),a''\right)\big\}\\
&=\max\big\{i(s'), \min_{a', a''\in\mathcal{A}}\tilde{b}\left(F(s',a'),a''\right)\big\}
\geq i(s')\stackrel{(s'\in\mathcal{G})}{=}1,
\end{align*}
which is a contradiction.  Hence $\mathcal{C}$ is safe.
\paragraph{Maximality of the CIS}
We finish by showing \textit{(ii)}. Assume (by contradiction) $\mathcal{C}$ is unreachable from outside. Assume furthermore $i(s)=0$ (if $i(s)=1$, this would mean $s\in\mathcal{G}$).\\
$\mathcal{C}$ reachable from outside means:
$$
\exists s\notin\mathcal{C}, \exists a\in\mathcal{A}: s'\triangleq F(s, a)\in\mathcal{C} \implies
$$
\begin{align*}
1&\stackrel{(s\notin\mathcal{C})}{=}\min_a\tilde{b}(s,a)\leq\tilde{b}(s, a) = \max\big\{i(s), \min_{a'}\tilde{b}(s',a')\big\}\\
&\stackrel{i(s)=0}{=} \min_{a'}\tilde{b}(s',a') \stackrel{(s'\in\mathcal{C})}{=} 0
\end{align*}
\end{proof}

    \subsection{Numerical experiments}\label{ap:hyperparams}
    \begin{table}[h]
    \centering
    \caption{Hyperparameters for inverted pendulum experiment}
    \begin{tabular}{l|c|c}
    &B2E (Ours) & SBE \\\hline\hline
    $\mathrm{dim}(\mathcal{S})$ & \multicolumn{2}{c}{3}                                                                                   
    \\\hline
    $|\mathcal{A}|$ & \multicolumn{2}{c}{5}                     \\\hline                                  
    NN hidden layers         & {[}256, 256{]}                                     & {[}256,256{]}                                      \\\hline
    NN activation            & {}Tanh{}                          & Tanh                                   \\\hline
    Learning rate\textsuperscript{\dag}             & \multicolumn{2}{c}{$(1-p)\times10^{-4}+p\times10^{-6}$}                                                   \\\hline
    Optimizer                    & \multicolumn{2}{c}{Adam}                                                                                \\\hline
    Discount $\gamma$        & N/A                                                & 0.9999                                             \\\hline
    Exploration factor       & 1                                                  & {$\max\{0.95\times 0.6^{p}, 0.05\}$}                                               \\\hline
    DDQN update              &     N/A                                               & Hard every 10 episodes                      \\\hline
    Buffer size               & \multicolumn{2}{c}{50000}\\\hline                 
    \end{tabular}
    \end{table}
    \textsuperscript{\dag}$p\triangleq$ progress, the fraction between the current episode and the total number of episodes.
    \paragraph{SBE safety critic}
    For Safety Bellman equation (SBE) \cite{fisac2019bridging}, the MDP at each step returns the signed distance to the unsafe set $h(s) = \frac{\pi}{2} -|\theta|$. The algorithm learns $q(s,a)$, and in principle any $(s,a)$ such that $q(s,a)\geq 0$ is safe.
    
    A more conservative safety-critic is one such that $q(s,a)\geq \eta$ for some $\eta>0$. When evaluating the learned models (Fig. \ref{fig:training-vs-dqn}, right) we consider different policies $\pi_\eta$, defined as the uniform-safe over the presumed safe actions (similar to our case):
    \begin{align*}
        \pi_\eta(a|s) = \begin{cases}
            0&\text{~if~}q_\eta(s,a)<\eta\\
            {1}/{\sum_{a'\in\mathcal{A}}\mathds{1}\{{q}^\theta(s,a')\geq\eta\}}&\text{~if~}{q}^\theta(s,a)\geq\eta
        \end{cases}\label{eq:uniform-safe-policy}
    \end{align*}
    
\end{document}

%% file: edits.tex
\usepackage{ifthen}
\newboolean{showcomments}
\setboolean{showcomments}{false}
\usepackage{todonotes}

\newboolean{arxiv}
\setboolean{arxiv}{false}

\definecolor{bleudefrance}{rgb}{0.19, 0.55, 0.91}
\definecolor{ao(english)}{rgb}{0.0, 0.5, 0.0}

\newcommand{\addcite}[0]{\ifthenelse{\boolean{showcomments}}
{\textcolor{purple}{(add cite(s)) }}{}}%

\newcommand{\addcites}[0]{\ifthenelse{\boolean{showcomments}}
{\textcolor{purple}{(add cite(s)) }}{}}%

\newcommand{\enrique}[1]{  \ifthenelse{\boolean{showcomments}}
{\todo[inline,color=bleudefrance]{Enrique: #1}}{}}

\newcommand{\emmargin}[1]{\ifthenelse{\boolean{showcomments}}{\marginpar{\color{bleudefrance}\tiny EM: #1}}{}}

\newcommand{\hmmargin}[1]{\ifthenelse{\boolean{showcomments}}{\marginpar{\color{orange}\tiny HM: #1}}{}}

\newcommand{\juan}[1]{  \ifthenelse{\boolean{showcomments}}
{\todo[inline,color=pink]{Juan: #1}}{}}

\newcommand{\hancheng}[1]{  \ifthenelse{\boolean{showcomments}}
{\todo[inline,color=orange]{Hancheng: #1}}{}}

\newcommand{\agustin}[1]{  \ifthenelse{\boolean{showcomments}}
{\todo[inline,color=purple!60!white]{Agustin: #1}}{}}

\newboolean{showedits}
\setboolean{showedits}{true}
\usepackage[markup=underlined]{changes}
\definechangesauthor[color=bleudefrance]{EM}
\newcommand{\aem}[1]{
\ifthenelse{\boolean{showedits}}
{\added[id=EM]{#1}}
{\!#1\hspace{-4.75pt}}
}
\newcommand{\repem}[2]{
\ifthenelse{\boolean{showedits}}
{\replaced[id=EM]{#1}{#2}}
{\!#1\hspace{-4.75pt}}
}
\newcommand{\dem}[1]{
\ifthenelse{\boolean{showedits}}
{\deleted[id=EM]{#1}}
{}
}

\newcommand{\supp}[0]{\operatorname{Supp}}



%% file: main.bbl
\begin{thebibliography}{10}

\bibitem{mnih2015human}
V.~Mnih, K.~Kavukcuoglu, D.~Silver, A.~A. Rusu, J.~Veness, M.~G. Bellemare, A.~Graves, M.~Riedmiller, A.~K. Fidjeland, G.~Ostrovski, {\em et~al.}, ``Human-level control through deep reinforcement learning,'' {\em nature}, vol.~518, no.~7540, pp.~529--533, 2015.

\bibitem{silver2016mastering}
D.~Silver, A.~Huang, C.~J. Maddison, A.~Guez, L.~Sifre, G.~Van Den~Driessche, J.~Schrittwieser, I.~Antonoglou, V.~Panneershelvam, M.~Lanctot, {\em et~al.}, ``Mastering the game of go with deep neural networks and tree search,'' {\em nature}, vol.~529, no.~7587, p.~484, 2016.

\bibitem{vinyals2017starcraft}
O.~Vinyals, T.~Ewalds, S.~Bartunov, P.~Georgiev, A.~S. Vezhnevets, M.~Yeo, A.~Makhzani, H.~K{\"u}ttler, J.~Agapiou, J.~Schrittwieser, {\em et~al.}, ``Starcraft ii: A new challenge for reinforcement learning,'' {\em arXiv preprint arXiv:1708.04782}, 2017.

\bibitem{nichols2019machine}
J.~A. Nichols, H.~W.~H. Chan, and M.~A. Baker, ``Machine learning: applications of artificial intelligence to imaging and diagnosis,'' {\em Biophysical reviews}, vol.~11, no.~1, pp.~111--118, 2019.

\bibitem{yu2021reinforcement}
C.~Yu, J.~Liu, S.~Nemati, and G.~Yin, ``Reinforcement learning in healthcare: A survey,'' {\em ACM Computing Surveys (CSUR)}, vol.~55, no.~1, pp.~1--36, 2021.

\bibitem{brunke2022safe}
L.~Brunke, M.~Greeff, A.~W. Hall, Z.~Yuan, S.~Zhou, J.~Panerati, and A.~P. Schoellig, ``Safe learning in robotics: From learning-based control to safe reinforcement learning,'' {\em Annual Review of Control, Robotics, and Autonomous Systems}, vol.~5, pp.~411--444, 2022.

\bibitem{gu2022review}
S.~Gu, L.~Yang, Y.~Du, G.~Chen, F.~Walter, J.~Wang, Y.~Yang, and A.~Knoll, ``A review of safe reinforcement learning: Methods, theory and applications,'' {\em arXiv preprint arXiv:2205.10330}, 2022.

\bibitem{paternain2022safe}
S.~Paternain, M.~Calvo-Fullana, L.~F. Chamon, and A.~Ribeiro, ``Safe policies for reinforcement learning via primal-dual methods,'' {\em IEEE Transactions on Automatic Control}, vol.~68, no.~3, pp.~1321--1336, 2022.

\bibitem{castellano2023learning}
A.~Castellano, H.~Min, J.~A. Bazerque, and E.~Mallada, ``Learning to act safely with limited exposure and almost sure certainty,'' {\em IEEE Transactions on Automatic Control}, vol.~68, no.~5, pp.~2979--2994, 2023.

\bibitem{chow2017risk}
Y.~Chow, M.~Ghavamzadeh, L.~Janson, and M.~Pavone, ``Risk-constrained reinforcement learning with percentile risk criteria,'' {\em The Journal of Machine Learning Research}, vol.~18, no.~1, pp.~6070--6120, 2017.

\bibitem{chen2023probabilistic}
W.~Chen, D.~Subramanian, and S.~Paternain, ``Probabilistic constraint for safety-critical reinforcement learning,'' {\em arXiv preprint arXiv:2306.17279}, 2023.

\bibitem{li2020robust}
S.~Li and O.~Bastani, ``Robust model predictive shielding for safe reinforcement learning with stochastic dynamics,'' in {\em 2020 IEEE International Conference on Robotics and Automation (ICRA)}, pp.~7166--7172, IEEE, 2020.

\bibitem{taylor2020learning}
A.~Taylor, A.~Singletary, Y.~Yue, and A.~Ames, ``Learning for safety-critical control with control barrier functions,'' in {\em Learning for Dynamics and Control}, pp.~708--717, PMLR, 2020.

\bibitem{robey2020learning}
A.~Robey, H.~Hu, L.~Lindemann, H.~Zhang, D.~V. Dimarogonas, S.~Tu, and N.~Matni, ``Learning control barrier functions from expert demonstrations,'' in {\em 2020 59th IEEE Conference on Decision and Control (CDC)}, pp.~3717--3724, IEEE, 2020.

\bibitem{castellano2022reinforcement}
A.~Castellano, H.~Min, E.~Mallada, and J.~A. Bazerque, ``Reinforcement learning with almost sure constraints,'' in {\em Learning for Dynamics and Control Conference}, pp.~559--570, PMLR, 2022.

\bibitem{bertsekas1972infinite}
D.~Bertsekas, ``Infinite time reachability of state-space regions by using feedback control,'' {\em IEEE Transactions on Automatic Control}, vol.~17, no.~5, pp.~604--613, 1972.

\bibitem{sontag2013mathematical}
E.~D. Sontag, {\em Mathematical control theory: deterministic finite dimensional systems}, vol.~6.
\newblock Springer Science \& Business Media, 2013.

\bibitem{bansal2017hamilton}
S.~Bansal, M.~Chen, S.~Herbert, and C.~J. Tomlin, ``Hamilton-jacobi reachability: A brief overview and recent advances,'' in {\em 2017 IEEE 56th Annual Conference on Decision and Control (CDC)}, pp.~2242--2253, IEEE, 2017.

\bibitem{gurriet2018towards}
T.~Gurriet, A.~Singletary, J.~Reher, L.~Ciarletta, E.~Feron, and A.~Ames, ``Towards a framework for realizable safety critical control through active set invariance,'' in {\em 2018 ACM/IEEE 9th International Conference on Cyber-Physical Systems (ICCPS)}, pp.~98--106, IEEE, 2018.

\bibitem{mitchell2007comparing}
I.~M. Mitchell, ``Comparing forward and backward reachability as tools for safety analysis,'' in {\em International Workshop on Hybrid Systems: Computation and Control}, pp.~428--443, Springer, 2007.

\bibitem{fisac2019bridging}
J.~F. Fisac, N.~F. Lugovoy, V.~Rubies-Royo, S.~Ghosh, and C.~J. Tomlin, ``Bridging hamilton-jacobi safety analysis and reinforcement learning,'' in {\em 2019 International Conference on Robotics and Automation (ICRA)}, pp.~8550--8556, 2019.

\bibitem{srinivasan2020learning}
K.~Srinivasan, B.~Eysenbach, S.~Ha, J.~Tan, and C.~Finn, ``Learning to be safe: Deep rl with a safety critic,'' {\em arXiv preprint arXiv:2010.14603}, 2020.

\bibitem{thananjeyan2021recovery}
B.~Thananjeyan, A.~Balakrishna, S.~Nair, M.~Luo, K.~Srinivasan, M.~Hwang, J.~E. Gonzalez, J.~Ibarz, C.~Finn, and K.~Goldberg, ``Recovery rl: Safe reinforcement learning with learned recovery zones,'' {\em IEEE Robotics and Automation Letters}, vol.~6, no.~3, pp.~4915--4922, 2021.

\bibitem{hsu2021safety}
K.-C. Hsu, V.~Rubies-Royo, C.~J. Tomlin, and J.~F. Fisac, ``Safety and liveness guarantees through reach-avoid reinforcement learning,'' {\em arXiv preprint arXiv:2112.12288}, 2021.

\bibitem{bertsekas2015dynamic}
D.~P. Bertsekas, ``Dynamic programming and optimal control 4th edition, volume ii,'' {\em Athena Scientific}, 2015.

\bibitem{schwartz93}
A.~Schwartz, ``A reinforcement learning method for maximizing undiscounted rewards,'' in {\em Proceedings of the tenth international conference on machine learning}, vol.~298, pp.~298--305, 1993.

\bibitem{girard2006efficient}
A.~Girard, C.~Le~Guernic, and O.~Maler, ``Efficient computation of reachable sets of linear time-invariant systems with inputs,'' in {\em Hybrid Systems: Computation and Control: 9th International Workshop, HSCC 2006, Santa Barbara, CA, USA, March 29-31, 2006. Proceedings 9}, pp.~257--271, Springer, 2006.

\bibitem{mitchell2005toolbox}
I.~M. Mitchell and J.~A. Templeton, ``A toolbox of hamilton-jacobi solvers for analysis of nondeterministic continuous and hybrid systems,'' in {\em International workshop on hybrid systems: computation and control}, pp.~480--494, Springer, 2005.

\bibitem{mitchell2005time}
I.~M. Mitchell, A.~M. Bayen, and C.~J. Tomlin, ``A time-dependent hamilton-jacobi formulation of reachable sets for continuous dynamic games,'' {\em IEEE Transactions on automatic control}, vol.~50, no.~7, pp.~947--957, 2005.

\bibitem{chen2021safe}
B.~Chen, J.~Francis, J.~Oh, E.~Nyberg, and S.~L. Herbert, ``Safe autonomous racing via approximate reachability on ego-vision,'' {\em arXiv preprint arXiv:2110.07699}, 2021.

\bibitem{blanchini1999set}
F.~Blanchini, ``Set invariance in control,'' {\em Automatica}, vol.~35, no.~11, pp.~1747--1767, 1999.

\bibitem{towers_gymnasium_2023}
M.~Towers, J.~K. Terry, A.~Kwiatkowski, J.~U. Balis, G.~d. Cola, T.~Deleu, M.~Goulão, A.~Kallinteris, A.~KG, M.~Krimmel, R.~Perez-Vicente, A.~Pierré, S.~Schulhoff, J.~J. Tai, A.~T.~J. Shen, and O.~G. Younis, ``Gymnasium,'' Mar. 2023.

\end{thebibliography}
